%% file: ms.tex
\documentclass[conference]{IEEEtran}

\pdfoutput=1
\input{packages}
\ifCLASSINFOpdf
\else
\fi
\ifCLASSOPTIONcompsoc
  \usepackage[caption=false,font=normalsize,labelfont=sf,textfont=sf]{subfig}
\else
  \usepackage[caption=false,font=footnotesize]{subfig}
\fi

\input{acronyms}

\input{macros}

\newcommand{\assoc}{A}
\newcommand{\assocspace}{\mathcal{A}}
\begin{document}
%

\input{head}
\maketitle

\begin{abstract}
\input{abstract}
\end{abstract}


%
\IEEEpeerreviewmaketitle

\input{introduction}

\input{background}

\input{notations}

\input{detections}

\input{tracking}

\input{results}

\input{conclusions}



\input{acknowledgment}


\IEEEtriggeratref{23}


\bibliographystyle{IEEEtran}
\bibliography{IEEEabrv,library.bib}
%



\end{document}

%% file: packages.tex
\usepackage{mathtools}
\usepackage{amssymb}
\usepackage{amsmath}

\usepackage{hyperref}
\usepackage{cleveref}

\usepackage[acronym,smallcaps]{glossaries} 

\usepackage[rgb]{xcolor}
\usepackage{pdfcomment}

\usepackage{placeins}

\usepackage{tabularx}
\usepackage{booktabs}
\usepackage{multirow}
\usepackage{threeparttable}

\usepackage{nth}

\usepackage{tikz}
\usepackage{siunitx}

%% file: acronyms.tex
\newacronym{cnn}{cnn}{convolutional neural network}
\newacronym{ppp}{ppp}{Poisson point process}
\newacronym{rfs}{rfs}{random finite set}
\newacronym{mb}{mb}{multi-Bernoulli}
\newacronym{mbm}{mbm}{multi-Bernoulli mixture}
\newacronym{stt}{stt}{single target tracking}
\newacronym{mot}{mot}{multi-object tracking}
\newacronym{iid}{iid}{independent and identically distributed}
\newacronym{pmbm}{pmbm}{Poisson multi-Bernoulli mixture}
\newacronym{pdf}{pdf}{probability density function}
\newacronym{jpda}{jpda}{joint probabilistic data association}
\newacronym{mht}{mht}{multiple hypothesis tracking}
\newacronym{phd}{phd}{probability hypothesis density}
\newacronym{cphd}{cphd}{cardinalized \textsc{phd}}
\newacronym{glmb}{glmb}{generalized labeled multi-Bernoulli}
\newacronym{rpn}{rpn}{region proposal network}
\newacronym{rcnn}{r-cnn}{region-based \textsc{cnn}}
\newacronym{yolo}{yolo}{you only look once}
\newacronym{mota}{mota}{\textsc{mot} accuracy}
\newacronym{motp}{motp}{\textsc{mot} precision}
\newacronym{f1}{f1}{F1 score}
\newacronym{pre}{pre}{precision}
\newacronym{rec}{rec}{recall}
\newacronym{far}{far}{false alarm rate}
\newacronym{iou}{iou}{intersection over union}
\newacronym{gt}{gt}{ground truth}
\newacronym{mt}{mt}{mostly tracked}
\newacronym{pt}{pt}{partially tracked}
\newacronym{ml}{ml}{mostly lost}
\newacronym{ids}{ids}{identity switches}
\newacronym{frag}{fr}{fragmentations}
\newacronym{fps}{fps}{frames per second}
\newacronym{cv}{cv}{constant velocity}
\newacronym{gospa}{gospa}{generalized optimal sub-pattern assignment}


%% file: macros.tex
\newcommand{\set}[1]{\boldsymbol{\mathbf{\MakeUppercase #1}}}
\newcommand{\card}[1]{\lvert#1\rvert}
\newcommand{\mm}[1]{\MakeUppercase #1}
\newcommand{\vv}[1]{\boldsymbol{\mathbf{\MakeLowercase #1}}}

\newcommand{\dsu}{\uplus}
\newcommand{\idxs}[1]{\mathbb{#1}}
\newcommand{\given}[2]{#1|#2}
\newcommand{\di}{\text{d}}
\newcommand{\men}[2]{\left[#1(\cdot)\right]^{#2}}
\newcommand{\ip}[2]{\left<#1;#2\right>}

\DeclareMathOperator*{\argmax}{arg\,max}

\newcommand{\tabitem}{\textbullet~~&}

\newcommand{\glsshort}[1]{\acronymfont{\glsentryshort{#1}}}



%% file: head.tex
\title{Mono-Camera 3D Multi-Object Tracking Using \\ Deep Learning Detections and PMBM Filtering}

\author{
  \IEEEauthorblockN{
    Samuel Scheidegger\IEEEauthorrefmark{1}\IEEEauthorrefmark{2},
    Joachim Benjaminsson\IEEEauthorrefmark{1}\IEEEauthorrefmark{2},
    Emil Rosenberg\IEEEauthorrefmark{2},
    Amrit Krishnan\IEEEauthorrefmark{1},
    Karl Granstr\"{o}m\IEEEauthorrefmark{2}
  }
  \IEEEauthorblockA{
    \IEEEauthorrefmark{1}Zenuity,
    \IEEEauthorrefmark{2}Department of Electrical Engineering, Chalmers University of Technology
    \\\IEEEauthorrefmark{1}\{firstname.lastname\}@zenuity.com, karl.granstrom@chalmers.se
  }
}

%% file: abstract.tex
Monocular cameras are one of the most commonly used sensors in the automotive industry for autonomous vehicles.
One major drawback using a monocular camera is that it only makes observations in the two dimensional image plane and can not directly measure the distance to objects. 
In this paper, we aim at filling this gap by developing a \glsentrylong{mot} algorithm that takes an image as input and produces trajectories of detected objects in a world coordinate system.
We solve this by using a deep neural network trained to detect and estimate the distance to objects from a single input image.
The detections from a sequence of images are fed in to a state-of-the art \glsentrylong{pmbm} tracking filter. 
The combination of the learned detector and the PMBM filter results in an algorithm that achieves 3D tracking using only mono-camera images as input.
The performance of the algorithm is evaluated both in 3D world coordinates, and 2D image coordinates, using the publicly available KITTI object tracking dataset.
The algorithm shows the ability to accurately track objects, correctly handle data associations, even when there is a big overlap of the objects in the image, and is one of the top performing algorithms on the KITTI object tracking benchmark. Furthermore, the algorithm is efficient, running on average close to 20 frames per second.


%% file: introduction.tex
\section{Introduction}

To enable a high level of automation in driving, it is necessary to accurately model the surrounding environment, a problem called environment perception.
Data from onboard sensors, such as cameras, radars and lidars, has to be processed to extract information about the environment needed to automatically and safely navigate the vehicle. 
For example, information about both the static environment, such as road boundaries and lane information, and the dynamic objects, like pedestrians and other vehicles, is of importance. The focus of this paper is the detection and tracking of multiple dynamic objects, specifically vehicles.

Dynamic objects are often modeled by state vectors, and are estimated over time using a \gls{mot} framework.
\Gls{mot} denotes the problem of, given a set of noisy measurements, estimating both the number of dynamic objects, and the state of each dynamic object.
Compared to the single object tracking problem, in addition to handling measurement noise and detection uncertainty, the \gls{mot} problem also has to resolve problems like object birth and object death\footnote{Object birth and object death is when an object first appears within, and departs from, the ego-vehicle's surveillance area, respectively.}; clutter detections\footnote{Clutter detections are false detections, i.e., detections not corresponding to an actual object.}; and unknown measurement origin.

A recent family of \gls{mot} algorithms are based on \glspl{rfs}~\cite{Mahler2007}.
The \gls{phd}~\cite{Mahler2003} filter, and the \gls{cphd}~\cite{Mahler2007b} filter, are two examples of moment approximations of the multi-object density.
The \gls{glmb}~\cite{Vo2013,Reuter2014} and the \gls{pmbm}~\cite{Williams2015,Garcia-Fernandez2017} filters are examples of \gls{mot} filters based on multi-object conjugate priors; these filters have been shown to outperform filters based on moment approximation. A recent comparison study published in~\cite{XiaGSGF:2017} has shown that the filters based on the \gls{pmbm} conjugate prior both achieves greater tracking performance, and has favourable computational cost compared to \gls{glmb}, hence we use the \gls{pmbm} filter in this work.

All of the aforementioned \gls{mot} algorithms takes sets of object estimates, or \textit{detections}, as their input. 
This implies that the raw sensor data, e.g., the images, should be pre-processed into detections.
The recent development of deep neural networks has lead to big improvement in fields of image processing.
Indeed, considerable improvements have be achieved for the object detection problem, see, e.g.,~\cite{Ren2015,Redmon2016}, which is is crucial to the tracking performance.


\Glspl{cnn} \cite{LeCun1998} have shown to vastly outperform previous methods in image processing for tasks such as classification, object detection and semantic segmentation.
\Glspl{cnn} make use of the spatial relation between neighbouring pixels in images, by processing data in a convolutional manner. Each layer in a \gls{cnn} consists of a filter bank with a number of convolutional kernels, where each element is a learnable parameter.

The most common approach for object detection using deep neural networks is \glspl{rcnn}. \Glspl{rcnn} are divided into two parts; a \gls{rpn}, followed by a box regression and classification network. The \gls{rpn} takes an image as input, and outputs a set of general object proposals, which are fed into the following classification and box regression network. The box regression and classification network will refine the size of the object and classify it into one of the object classes. This type of deep neural network structure is used in, e.g., Fast \gls{rcnn}~\cite{Girshick2015}, and later in the improved Faster \gls{rcnn}~\cite{Ren2015}. Another approach to the object detection problem is \gls{yolo}~\cite{Redmon2016}. Here, the region proposal step is omitted, and the box regression and classification are applied directly on the entire image.

In the automotive industry, monocular camera is a well studied and commonly used type of sensors for developing autonomous driving systems.
A monocular camera is a mapping between 3D world coordinates and 2D image coordinates~\cite{RichardHartley2004} where, in contrary to, e.g., radars and lidars, distance information is lost.
However, to achieve a high level of automation, tracking in the image plane is not adequate.
Instead, we need to track objects in world coordinates in order to obtain the relative pose between the ego vehicle the detected objects, information that is crucial for automatic decision making and control.
We refer to this as 3D tracking.

Previous work on object tracking using monocular camera data is restricted to tracking in the image-plane, see, e.g.,~\cite{Choi2016,He2016,Xiang2015}, for some recent work. The main contribution of this paper is a multi-vehicle 3D tracking algorithm, that takes as input mono camera data, and outputs vehicle estimates in world coordinates. The proposed \gls{mot} algorithm is evaluated using the image sequences from the publicly available KITTI tracking dataset~\cite{Geiger2012CVPR}, and the results show that accurate 3D tracking is achieved.

The presented 3D tracking filter has two main components: a detector and an object tracking filter.
The detector is a deep neural network trained to from an input image not only extract a 2D bounding box for each detected object, but also to estimate the distance from the camera to the object.
This is achieved by using object annotations in lidar data during the learning of the network parameters.
The object tracking filter is a state-of-the-art \gls{pmbm} object tracking filter~\cite{Williams2015,Garcia-Fernandez2017} that processes the detections and outputs estimates.
The tracking filter is computationally efficient, and handles both false detections and missed detections.
For each object, a position, as well as kinematical properties such as velocity, are estimated.

The paper is structured as follows. In Section~\ref{sec:ProblemFormulationAlgorithmOverview}, we give a problem formulation and present an overview of the algorithm. In Section~\ref{sec:Detections} we present the object detection, and in Section~\ref{sec:Tracking} we present object tracking.
The results of an experimental evaluation using data sequences from the KITTI dataset are presented in Section~\ref{sec:results}%
, and the paper is concluded in Section~\ref{sec:Conclusion}.

%% file: background.tex
\section{Problem formulation and algorithm overview}
\label{sec:ProblemFormulationAlgorithmOverview}
%

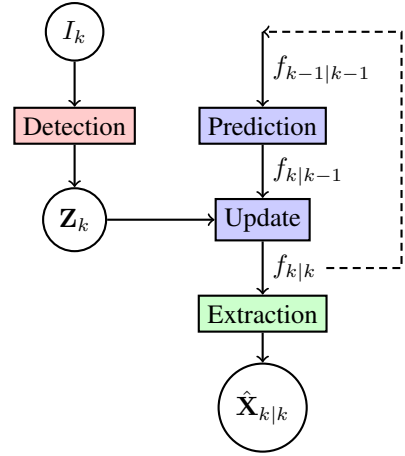
\begin{figure}[t]
    \centering

    \input{images/system.tikz}
    \caption{Algorithm overview. The \gls{mot} algorithm has two modules, detection (left, red) and tracking (right, blue/green). The tracking modules consists of a recursive tracking filter (prediction+update) and an object extraction.}
    \label{fig:system}
\end{figure}

The KITTI object tracking dataset~\cite{Geiger2012CVPR} contains data from multiple sensors, e.g., four cameras and a lidar sensor. The data from such sensors can be used for environment perception, i.e., tracking of moving objects and mapping of the stationary environment. In this work we focus on data from a forward looking camera, with the objective to track the other vehicles that are in the environment.

Each vehicle is represented by a state vector $\vv{x}$ that contains the relevant information about the object. For 3D object tracking, the following state vector is used,
\begin{align}
	\vv{x} = \begin{bmatrix} x & y & z & v_x & v_y & v_z & w & h \end{bmatrix}^{T},
\end{align}
where $(x,y,z)$ is the 3D position in world coordinates, $(v_x , v_y , v_z)$ is the corresponding velocity, and $(w,h)$ is the width and height of the object's bounding box in the camera image. The position and velocity describes the tracked object's properties of interest; the width and height of the bounding box are used for evaluation analogue to the KITTI object tracking benchmark~\cite{Geiger2012CVPR}.


The number of vehicles in the environment is not known, and changes with time, so the task is to estimate both the number of vehicles, as well as each vehicle's state. The vehicles at time step $k$ are represented by a set $\set{X}_{k}$ that contains the state vectors of all vehicles that are present in the vicinity of the ego-vehicle. The set of vehicles $\set{X}_{k}$ is modeled as a Random Finite Set (\gls{rfs}) \cite{Mahler2007}.  That is, the number of objects, or the cardinality of the set, is modeled as a time varying discrete random variable and each object's state is a multivariate random variable.

The problem addressed in this paper is the processing of the sequence of images $\mm{I}_k$ into a sequence of estimates $\hat{\set{X}}_{\given{k}{k}}$ of the set of vehicles,
\begin{align}
	I_{0},I_{1},\ldots,I_{k} \qquad \Rightarrow \qquad \hat{\set{X}}_{0|0},\hat{\set{X}}_{1|1},\ldots,\hat{\set{X}}_{k|k},
\end{align}
where the sub-indices denote time. In other words, we wish to process the image sequence to gain information at each time step about the number of vehicles (the cardinality of the set $\set{X}$), and the state of each vehicle. 
%
%
The proposed \gls{mot} algorithm has two main parts: object detection, and object tracking; an illustration of the algorithm is given in~\cref{fig:system}.

In the detection module, each image is processed to output a set of object detections $\set{Z}_{k}$,
\begin{align}
	I_{k} \quad \xRightarrow{\text{Detection}} \quad \set{Z}_{k}.
\end{align}
The set of detections $\set{Z}_{k}$, where each  $\vv{z}_k^i\in\set{Z}_{k}$ is an estimated object, is also modeled as a \gls{rfs}. The detection is based on a \gls{cnn}, which is presented in detail in Section~\ref{sec:Detections}.

The tracking module takes the image detections as input and outputs an object set estimate; it has three parts: prediction, update, and extraction. Together, the prediction and the update constitute a tracking filter that recursively estimates a multi-object set density,
\begin{align}
	\set{Z}_{k} \quad \xRightarrow{\text{PMBM filter}} \quad f_{k|k}(\set{X}_{k} | \set{Z}^k),
\end{align}
where $\set{Z}^k$ denotes all measurement sets up to time step $k$, $\{\set{Z}_t\}_{t\in(0,k)}$. Specifically, in this work we estimate a \gls{pmbm} density \cite{Williams2015}. The Chapman-Kolmogorov prediction
\begin{subequations}
    \begin{multline}
      f_{\given{k}{k-1}}(\given{\set{X}_k}{\set{Z}^{k-1}}) \\
      =\int g(\given{\set{X}_k}{\set{X}_{k-1}})f_{\given{k-1}{k-1}}(\given{\set{X}_{k-1}}{\set{Z}^{k-1}})\delta\set{X}_{k-1},
    \end{multline}
predicts the \gls{pmbm} density to the next time step using the multi-object motion model $g(\given{\set{X}_{k+1}}{\set{X}_k})$. We use the standard multi-object motion model~\cite{Mahler2007}, meaning that $g(\given{\cdot}{\cdot})$ models a Markovian process for objects that remain in the field of view, combined a \gls{ppp} birth process.

Using the set of detections $\set{Z}_{k}$ and the multi-object measurement model $h(\given{\set{Z}_k}{\set{X}_k})$, the updated \gls{pmbm} density is computed using the Bayes update
  \begin{equation}
    f_{\given{k}{k}}(\given{\set{X}_k}{\set{Z}^k})=
    \frac{h(\given{\set{Z}_k}{\set{X}_k})f_{\given{k}{k-1}}(\given{\set{X}_k}{\set{Z}^{k-1}})}
    {\int h(\given{\set{Z}_k}{\set{X}_k})f_{\given{k}{k-1}}(\given{\set{X}_k}{\set{Z}^{k-1})\delta\set{X}_k}},
  \end{equation}%
  \label{eq:MultiObjectBayesFilter}%
\end{subequations}%
We use the standard multi-object measurement model~\cite{Mahler2007}, in which $h(\given{\set{Z}_k}{\set{X}_k})$ models noisy measurement with detection uncertainty, combined with \gls{ppp} clutter.

The final part of the tracking is the object extraction, where object estimates are extracted from the \gls{pmbm} density,
\begin{align}
	f_{k|k}(\set{X}_{k} | \set{Z}^k) \quad \xRightarrow{\text{Extraction}} \quad \hat{\set{X}}_{k|k}
\end{align}
The tracking is described further in Section~\ref{sec:Tracking}. The integrals in \eqref{eq:MultiObjectBayesFilter} are set-integrals, defined in~\cite{Mahler2007}.



%% file: images/system.tikz
\begin{tikzpicture}


\node[circle,draw,thick] at (-0.5,4.75) (I){$I_k$};
\node[rectangle,draw,thick,fill=red!20!white] at (-0.5,3.5) (Det){Detection};
\node[circle,draw,thick] at (-0.5,2.25) (Z){$\set{Z}_k$};

\draw[->,thick] (I) -- (Det);
\draw[->,thick] (Det) -- (Z);

\coordinate (next) at (2,4.75);
\node[rectangle,draw,thick,fill=blue!20!white] at (2,3.5) (Pre){Prediction};
\node[rectangle,draw,thick,fill=blue!20!white] at (2,2.25) (Up){Update};
\node[rectangle,draw,thick,fill=green!20!white] at (2,1) (Ex){Extraction};
\node[circle,draw,thick] at (2,-0.25) (X){$\hat{\set{X}}_{\given{k}{k}}$};

\draw[->,thick] (Z) -- (Up);
\draw[->,thick] (Pre) -- (Up) node [midway,right] (fkkm) {$f_{\given{k}{k-1}}$};
\draw[->,thick] (Up) -- (Ex) node [midway,right] (fkk) {$f_{\given{k}{k}}$};
\draw[->,thick] (next) -- (Pre) node [midway,right] (fkmkm) {$f_{\given{k-1}{k-1}}$};
\draw[->,densely dashed,thick] (fkk.east) -- ++(1,0) |- (next);
\draw[->,thick] (Ex) -- (X);



\end{tikzpicture}


%% file: notations.tex
\begin{table}[t]\caption{Table of Notation}
  \begin{center}
    \begin{tabularx}{\columnwidth}{r@{}X}
      \toprule

      \tabitem Minor non-bold letter, e.g., $a$, $b$, $\gamma$, denote
      scalars. \\
      \tabitem Minor bold letters, e.g.,  $\vv{x}$, $\vv{z}$, $\vv{\xi}$, denote vectors. \\
      \tabitem Capital non-bold letters , e.g., $\mm{M}$, $\mm{F}$,
      $\mm{H}$, denote matrices. \\
      \tabitem Capital bold letters, e.g., $\set{X}$, $\set{Y}$, $\set{Z}$, denote sets. \\
      \tabitem $\card{\set{X}}$ denotes the cardinality of set $\set{X}$,
      i.e., the number of elements in $\set{X}$. \\
      \tabitem $\dsu$ denotes disjoint set union, i.e.,
      $\set{X}\dsu\set{Y}=\set{Z}$ means $\set{X}\cup\set{Y}=\set{Z}$ and
      $\set{X}\cap\set{Y}=\emptyset$. \\
      \tabitem $\men{h}{\set{X}}=\prod_{\vv{x}\in\set{X}}h(\vv{x})$
      and $\men{h}{\emptyset}=1$ by definition.\\
      \tabitem $\ip{a}{b}=\int a(x)b(x)\di x$, the inner product of $a(x)$ and $b(x)$. \\

      \bottomrule
    \end{tabularx}
  \end{center}
\end{table}

%% file: detections.tex
\section{Object Detection}
\label{sec:Detections}

In this section, we describe how deep learning, see, e.g., \cite{Lecun2015}, is used to process the images $\{I_{t}\}_{t=0}^{k}$ to output sets of detections $\{\set{Z}_{t}\}_{t=0}^{k}$. For an image $I_{t}$ with a corresponding set of detections $\set{Z}_{t}$, each detection $\vv{z} \in \set{Z}_{t}$ consists of a 2D bounding box and a distance from the camera center to the center of the detected object,
\begin{align}
  \vv{z} = \begin{bmatrix} x_{min} & y_{min} & x_{max} & y_{max} & d \end{bmatrix}^{T},
\end{align}
where $(x_{min},y_{min})$ and $(x_{max},y_{max})$ are the pixel positions of the top left and bottom right corner of the bounding box,
respectively, and $d$ is the distance from the camera to the object. The bounding box encloses the object in the image. Using this information, the angle from the camera center to the center of the detected object can be inferred. This, together with the camera-to-object-distance $d$, allows the camera to be transformed into a range/bearing sensor, which is suitable for object tracking in 3D world coordinates.

The object detection is implemented using a improved version of the network developed in~\cite{Krishnan2016}.
The network can be divided into two parts; the first part can be viewed as a feature extractor, and the second part consists of three parallel output headers.
The feature extractor is identical to the DRN-C-26~\cite{Yu2017} network, with the exception that the last two classification layers have been removed.
The last two layers are structured for the original classification task of DRN-C-26, which is not suitable in this work.

To represent objects using a bounding box and its distance, the network has three different types of output: classification score, bounding box and distance.
Each header in the network has two $1\times 1$ convolutional layers and finally a sub-pixel convolutional layer~\cite{Shi2016}, upscaling the output to 1/4th of the input image resolution.
The bounding box header has 4 output channels, representing the top left and bottom right corner of the bounding box, the distance header has one output channel, representing the distance to the object, and the classification header has an additional softmax function and represents the different class scores using one-hot encoding, i.e., one output channel for each class, where each channel represents the score for each class, respectively.
For each pixel in the output layer there will be an estimated bounding box, i.e., there can be more than one bounding box per object.
To address this, Soft-NMS~\cite{Bodla2017} is applied.
In this step, the box with the highest classification score is selected and the score of boxes intersecting the selected box are decayed according to a function of the \gls{iou}.
This process is repeated until the classification score of all remaining boxes are below a manually chosen threshold.

The feature extractor is pre-trained on ImageNet~\cite{JiaDeng2009} and the full network is fine-tuned using annotated object labels from the KITTI object data set~\cite{Geiger2012CVPR}.
The network is tuned using stochastic gradient descent with momentum. The task of classification used a cross entropy loss function while bounding box regression and distance estimation used a smooth L1 loss function~\cite{Girshick2015}.


%% file: tracking.tex
\section{Object tracking}
\label{sec:Tracking}
To associate objects between consecutive frames and filter the object detections from the neural network, a \gls{pmbm} tracking filter is applied.
Both the set of objects $\set{X}_{k}$ and the set of image detections $\set{Z}_{k}$ are modeled as \glspl{rfs}. The purpose of the tracking module is to process the sequence of detection sets, and output a sequence of estimates $\hat{\set{X}}_{\given{k}{k}}$ of the true set of objects. We achieve this by using a \gls{pmbm} filter to estimate the multi-object density $f_{\given{k}{k}}(\given{\set{X}_{k}}{\set{Z}_{k}})$, and to extract estimates from this density.

In this section, we first present some necessary \gls{rfs} background, and the standard point object models that are used to model both the object motion, as well as the detection process. Then, we present the \gls{pmbm} filter.


\subsection{\gls{rfs} background}
In this work, two types of \glspl{rfs} are important: the \gls{ppp} and the Bernoulli process. A general introduction to \gls{rfs} is given in, e.g.,~\cite{Mahler2007}.

\subsubsection{\Glsentrylong{ppp}}
A \gls{ppp} is a type of \gls{rfs} where the cardinality is Poisson
distributed and all elements are \gls{iid}.
A \gls{ppp} can be parametrized by an intensity function, $D(\vv{x})$, defined as
\begin{equation}
  D(\vv{x})=\mu f(\vv{x}).
\end{equation}
The intensity function has two parameters, the Poisson rate $\mu>0$ and the spatial distribution $f(\vv{x})$.
The expected number of set members in a \gls{ppp} S is $\int_{\vv{x}\in S}D(\vv{x})\di\vv{x}$.

The \gls{ppp} density is
\begin{equation}
  f(\set{X})=e^{-\ip {D(\vv{x})}{1}}\prod_{\vv{x}\in\set{X}}D(\vv{x})=e^{-\mu}\prod_{\vv{x}\in\set{X}}\mu f(\vv{x}).
  \label{eq:poisson}
\end{equation}
The \glspl{ppp} are used to model object birth, undetected
objects and clutter measurements.

\subsubsection{Bernoulli process}
A Bernoulli \gls{rfs} is a \gls{rfs} that with the probability $r$ contains a single element with the \gls{pdf} $f(\vv{x})$, and with the probability $1-r$ is empty:
\begin{equation}
  f(\set{X})=
  \begin{cases}
    1-r, & \set{X}=\emptyset \\
    rf(\vv{x}), & \set{X}=\{\vv{x}\} \\
    0, & \card{X} > 1
  \end{cases}.
  \label{eq:bernoulli_set_densities}
\end{equation}
It is suitable to use a Bernoulli \gls{rfs} to model objects in a \gls{mot} problem, since it both models the object's probability of existence $r$, and uncertainty in its state $\vv{x}$.

In \gls{mot}, the objects are typically assumed to be
independent~\cite{Williams2015}.
%
%
The disjoint union of a fixed number of independent Bernoulli \glspl{rfs},
$\set{X}=\dsu_{i\in\idxs{I}}\set{X}^i$, where $\idxs{I}$
is an index set, is a \gls{mb} \gls{rfs}.
The parameters $\{r^i,f^i(\cdot)\}_{i\in\idxs{I}}$
defines the \gls{mb} distribution.

A \gls{mbm} density is a normalized, weighted sum of \gls{mb} densities.
The \gls{mbm} density is entirely defined by $\{w^j,\{r^{j,i},f^{j,i}(\cdot)\}_{i\in\idxs{I}^j}\}_{j\in\idxs{J}}$, where $\idxs{J}$ is an index set for the \glspl{mb} in the \gls{mbm}, $w^j$ is the probability of the $j$th \gls{mb}, and $\idxs{I}^j$ is the index set for the Bernoulli distributions.
In a \gls{mot} problem, the different \glspl{mb} typically corresponds
to different data association sequences.

\subsection{Standard models}
Here we present the details of the standard measurement and motion models, under Gaussian assumptions.
\subsubsection{Measurement model}
\label{sec:standardMeasModel}
Let $\vv{x}_{k}^{i}$ be the state of the $i$th vehicle at the $k$th time step.
At time step $k$, given a set of objects
$\set{X}_k=\{\vv{x}_k^i\}_{i\in\idxs{I}}$, the set of measurements is
$\set{Z}_k=(\dsu_{i\in\idxs{I}}\set{W}_k^i)\dsu\set{K}_k$, where
$\set{W}_k^i$ denotes the set of object generated measurements from
the $i$th object, $\idxs{I}$ is an index set and $\set{K}_k$ denotes
the set of clutter measurements.
The set $\set{K}_k$ is modeled as a \gls{ppp}
with the intensity $\kappa(\vv{z})=\lambda c(\vv{z})$, where $\lambda$ is the Poisson rate and the spatial distribution $c(\vv{z})$ is assumed to be uniform.

Assuming an object is correctly detected with probability of detection $p_{\rm D}$. If the object is detected, the measurement
$\vv{z}\in\set{W}_k^i$ has \gls{pdf}
$\phi_{\vv{z}}(\vv{x}_k^i) = \mathcal{N}(\vv{z} ; a(\vv{x}_k^i), R)$, where $a(\vv{x}_k^i)$ is a camera measurement model.
The resulting measurement likelihood is%
\begin{equation}
  \ell_{\set{Z}}(\vv{x}) = p(\given{\set{Z}}{\vv{x}})=\begin{cases}
    1-p_{\rm D}, & \set{Z}=\emptyset \\
    p_{\rm D} \phi_{\vv{z}}(\vv{x}), & \set{Z}=\{\vv{z}\} \\
    0, & \card{\set{Z}}>1
  \end{cases}.
  \label{eq:mptdb}
\end{equation}
\label{eq:mptd}%
As can be seen in \cref{eq:mptdb}, if multiple measurements are associated to one object this will have zero likelihood. This is a standard point object assumption, see, e.g., \cite{Mahler2007}.

Because of the unknown measurement origin\footnote{An inherent property of \gls{mot} is that it is unknown which measurements are from object and which are clutter, and among the object generated measurements, it is unknown which object generated which measurement. Hence, the update must handle this uncertainty.}, it is necessary to discuss data association. Let the measurements in the set $\set{Z}$ be indexed by $m\in\mathbb{M}$,
\begin{align}
	\set{Z} = \left\{\vv{z}^m\right\}_{m\in\mathbb{M}}, 
\end{align}
and let $\assocspace^{j}$ be the space of all data associations $\assoc$ for the $j$th predicted global hypothesis, i.e., the $j$th predicted \gls{mb}. A data association $\assoc\in\assocspace^{j}$ is an assignment of each measurement in $\set{Z}$ to a source, either to the \emph{background} (clutter or new object) or to one of the existing objects indexed by $i\in\mathbb{I}^{j}$. Note that $\mathbb{M} \cap \mathbb{I}^{j} = \emptyset$ for all $j$. The space of all data associations for the $j$th hypothesis is $\assocspace^{j} = \mathcal{P}(\mathbb{M} \cup \mathbb{I}^{j})$, i.e., a data association $\assoc\in\assocspace^{j}$ is a partition of $\mathbb{M} \cup \mathbb{I}^{j}$ into non-empty disjoint subsets $C\in\assoc$, called index cells\footnote{For example, let $\mathbb{M}=\left(m_1,m_2,m_3\right)$ and $\mathbb{I}=\left( i_{1} , i_{2} \right)$, i.e., three measurements and two objects. One valid partition of $\mathbb{M}\cap\mathbb{I}$, i.e., one of the possible associations, has the following four cells $\{m_1\},\{m_2,i_{1}\},\{m_3\},\{i_{2}\}$. The meaning of this is that measurement $m_2$ is associated to object $i_{1}$, object $i_{2}$ is not detected, and measurements $m_1$ and $m_3$ are not associated to any previously detected object, i.e., measurements $m_1$ and $m_3$ are either clutter or from new objects.}.

Due to the standard \gls{mot} assumption that the objects generate measurements independent of each other, an index cell contains at most one object index and at most one measurement index, i.e., $| C\cap\mathbb{I}^{j} | \leq 1$ and $| C\cap\mathbb{M} | \leq 1$ for all $C\in\assoc$. Any association in which there is at least one cell, with at least two object indices and/or at least two measurement indices, will have zero likelihood because this violates the independence assumption and the point object assumption, respectively. If  the index cell $C$ contains an object index, then let $i_C$ denote the corresponding object index, and if the index cell $C$ contains a measurement index, then let $m_C$ denote the corresponding measurement index.

\subsubsection{Standard dynamic model}
\label{sec:StandardDynamicModel}
The existing objects---both the detected and the undetected---survive from time step $k$ to time step $k+1$ with probability of survival $p_{\rm S}$. The objects evolve independently according to a Markov process with Gaussian transition density $g(\vv{x}_{k+1} | \vv{x}_{k}) = \mathcal{N}(\vv{x}_{k+1} ; b(\vv{x}_{k}), Q)$, where $b(\vv{\cdot})$ is a \gls{cv} motion model. New objects appear independently of the objects that already exist. The object birth is assumed to be a \gls{ppp} with intensity $D_{k+1}^{b}(\vv{x})$, defined in~\cref{eq:poisson}.

\subsection{\gls{pmbm} filter}
In this section, the time indexing has been omitted for notational
simplicity.
The \gls{pmbm} filter is a combination of two \glspl{rfs}, a \gls{ppp}
to model the objects that exist at the current time step, but have
not yet been detected and a \gls{mbm} to model the objects that have
been detected previously at least once.
The set of objects can be divided into two disjoint subsets,
$\set{X}=\set{X}^d\dsu\set{X}^u$, where $\set{X}^d$ is the set of
detected objects and $\set{X}^u$ is the set of undetected objects.
The \gls{pmbm} density can be expressed as
\begin{subequations}
  \begin{align}
    f(\set{X}) & = \sum_{\set{X}^u\dsu\set{X}^d=\set{X}}f^u(\set{X}^u)\sum_{j\in\idxs{J}}w^jf^j(\set{X}^d), \\
    f^u(\set{X}^u) & = e^{-\ip{D^u(\vv{x})}{1}} \men{D^u}{\set{X}^u}, \label{eq:undetected_objects} \\
    f^j(\set{X}^d) & = \sum_{\dsu_{i\in\idxs{I}^i}\set{X}^i=\set{X}^d}\prod_{i\in\idxs{I}^j}f^{j,i}(\set{X}^i), \label{eq:object_mb}
  \end{align}%
  \label{eq:PMBMdensity}%
\end{subequations}%
where
\begin{itemize}
\item $f^u(\cdot)$ is the \gls{ppp} density for the set of undetected
  objects $\set{X}^u$, where $D^u(\cdot)$ is its intensity.
\item $\idxs{J}$ is an index set of \gls{mbm} components. There are $\card{\idxs{J}}$ \glspl{mb}, where each \gls{mb} corresponds to a unique global data association hypothesis. The probability of each component in the \gls{mbm} is denoted as $w^j$.
\item For every component $j$ in the \gls{mbm}, there is an index set $\idxs{I}^j$, where each index $i$ corresponds to a potentially detected object $\set{X}^i$. 
\item $f^{j,i}(\cdot)$ are Bernoulli set densities, defined in \cref{eq:bernoulli_set_densities}. Each \gls{mb} corresponds to a potentially detected object with a probability of existence and a state \gls{pdf}.
\end{itemize}
The \gls{pmbm} density in~\cref{eq:PMBMdensity} is defined by the involved parameters,
\begin{align}
	D_{}^{u} , \{ (w_{}^{j}, \{(r_{}^{j,i},f_{}^{j,i})\}_{i\in\mathbb{I}_{}^{j}})\}_{j\in\mathbb{J}_{}}.
\end{align}
Further, the \gls{pmbm} density is an \gls{mot} conjugate prior \cite{Williams2015}, meaning that for the standard point object models (Sections~\ref{sec:standardMeasModel} and \ref{sec:StandardDynamicModel}), the prediction and update in~\cref{eq:MultiObjectBayesFilter} both result in \gls{pmbm} densities. It follows that the \gls{pmbm} filter propagates the multi-object density by propagating the set of parameters.

In this work, we assume that the birth intensity $D^b$ is a non-normalized Gaussian mixture. It follows from this assumption that the undetected intensity $D^u$ is also a non-normalized Gaussian mixture, and all Bernoulli densities $f^{j,i}$ are Gaussian densities. Below, we present the parameters that result from the prediction and the update, and we present a simple method for extracting target estimates from the set of parameters. To compute the predicted and updated Gaussian parameters, we use the UKF prediction and update, respectively, see, e.g., \cite[Ch. 5]{Sarkka2013}.

\subsubsection{Prediction}
Given a posterior \gls{pmbm} density with parameters
\begin{align}
	D_{}^{u} , \{ (w_{}^{j}, \{(r_{}^{j,i},f_{}^{j,i})\}_{i\in\mathbb{I}_{}^{j}})\}_{j\in\mathbb{J}_{}} ,
\end{align}
and the standard dynamic model (Section \ref{sec:StandardDynamicModel}), the predicted density is a \gls{pmbm} density with parameters
\begin{subequations}
\begin{align}
	D_{+}^{u} , \{ (w_{+}^{j}, \{(r_{+}^{j,i},f_{+}^{j,i})\}_{i\in\mathbb{I}^{j}})\}_{j\in\mathbb{J}},
\end{align}
where

\begin{align}%
	D_{+}^{u} (\vv{x}) & = D^{b}(\vv{x}) + p_{\rm S} \ip{D_{}^{u}}{ g} , \label{eq:PredictionUndetected} \\
	 r_{+}^{j,i} & = p_{\rm S} r^{j,i} , \label{eq:PredictedBernoulliProbEx} \\
	 f_{+}^{j,i}(\vv{x}) & = \ip{ f_{}^{j,i}}{g} , \label{eq:PredictedBernoulliSpatialDistribution}
\end{align}%
\label{eq:PredictedParameters}%
\end{subequations}%
and $w_{+}^{j} = w_{}^{j} $. For Gaussian mixture intensity $D^u$, and Gaussian densities $f^{j,i}$, the predictions $\ip{\cdot}{g}$ in~\cref{eq:PredictedParameters} are easily computed using the UKF prediction, see, e.g.,~\cite[Ch. 5]{Sarkka2013}.

\subsubsection{Update}
Given a prior \gls{pmbm} density with parameters
\begin{align}
	D_{+}^{u} , \{ (w_{+}^{j}, \{(r_{+}^{j,i},f_{+}^{j,i})\}_{i\in\mathbb{I}_{+}^{j}})\}_{j\in\mathbb{J}_{+}},
\end{align}
a set of measurements $\set{Z}$, and the standard measurement model (Section~\ref{sec:standardMeasModel}), the updated density is a \gls{pmbm} density
\begin{subequations}
\begin{align}
	f(\set{X}|\set{Z}) & = \sum_{\set{X}^{u} \uplus \set{X}^{d} = \set{X}} f^{u}(\set{X}^{u}) \sum_{j\in\mathbb{J}_{+}} \sum_{\assoc\in\assocspace^{j}} w_{\assoc}^{j} 	f_{\assoc}^{j}(\set{X}^{d}) , \\
	f^{u}(\set{X}^{u}) & = e^{-\ip{D^{u}}{1}} \prod_{\vv{x}\in\set{X}^{u}} D^{u}(\vv{x}) , \\
	f_{\assoc}^{j}(\set{X}^{d}) & = \sum_{ \uplus_{C\in\assoc}\set{X}^{C} = \set{X} } \prod_{C\in\assoc} f_{C}^{j}(\set{X}^{C}) ,
\end{align}%
\label{eq:UpdatedPMBMdensity}%
where the weights are
\begin{align}
	& w_{\assoc}^{j} = \frac{w_{+}^{j} \prod_{C\in\assoc} \mathcal{L}_C }{ \sum_{j'\in\mathbb{J}} \sum_{\assoc'\in\assocspace^{j'}} w_{+}^{j'} \prod_{C'\in\assoc'} \mathcal{L}_{C'} } , \\
	& \mathcal{L}_{C} = \left\{ \begin{array}{cl}
			\text{\footnotesize $\kappa + p_{\rm D} \ip{ D_{+}^{u} }{ \phi_{\vv{z}^{m_{C}}} }$}  & \text{\footnotesize if $C\cap\mathbb{I}^{j} = \emptyset, C\cap\mathbb{M} \neq \emptyset,$} \\
			\text{\footnotesize $1- r_{+}^{j,i_{C}} p_{\rm D} $}     & \text{\footnotesize if $C\cap\mathbb{I}^{j}\neq\emptyset, C\cap\mathbb{M}=\emptyset,$} \\
			\text{\footnotesize $r_{+}^{j,i_{C}} p_{\rm D} \ip{f_{+}^{j,i_{C}}}{ \phi_{\vv{z}^{m_{C}}}}$}                  & \text{\footnotesize if $C\cap\mathbb{I}^{j}\neq\emptyset, C\cap\mathbb{M}\neq\emptyset ,$}
		\end{array} \right.
\end{align}
the densities $f_{C}^{j}(\set{X})$ are Bernoulli densities with parameters
\begin{align}
	& r_{C}^{j} = \left\{ \begin{array}{cl}
			\frac{ p_{\rm D} \ip{ D_{+}^{u} }{ \phi_{\vv{z}^{m_{C}}} } }{\kappa+ p_{\rm D}\ip{ D_{+}^{u} }{ \phi_{\vv{z}^{m_{C}}} }} & \text{\footnotesize if $C\cap\mathbb{I}^{j} = \emptyset, C\cap\mathbb{M} \neq \emptyset$,} \\
			\frac{r_{+}^{j,i_{C}} (1 - p_{\rm D}) }{1-r_{+}^{j,i_{C}} p_{\rm D} } & \text{\footnotesize if $C\cap\mathbb{I}^{j}\neq\emptyset, C\cap\mathbb{M}=\emptyset$,} \\
			1 & \text{\footnotesize if $C\cap\mathbb{I}^{j}\neq\emptyset, C\cap\mathbb{M}\neq\emptyset $,}
		\end{array} \right. \label{eq:NewBernoulliProbEx}\\
	& f_{C}^{j}(\vv{x}) = \left\{ \begin{array}{cl}
			\frac{ \phi_{\vv{z}^{m_{C}}}(\vv{x}) D_{+}^{u}(\vv{x}) }{ \ip{ D_{+}^{u} }{ \phi_{\vv{z}^{m_{C}}} } } & \text{\footnotesize if $C\cap\mathbb{I}^{j} = \emptyset, C\cap\mathbb{M}\neq\emptyset$,} \\
			\text{\footnotesize $f_{+}^{j,i_{C}}(\vv{x})$} & \text{\footnotesize if $C\cap\mathbb{I}^{j}\neq\emptyset, C\cap\mathbb{M}=\emptyset$,} \\
			\frac{ \phi_{\vv{z}^{m_{C}}}(\vv{x}) f_{+}^{j,i_{C}} (\vv{x})}{\ip{ f_{+}^{j,i_{C}} }{ \phi_{\vv{z}^{m_{C}}} }} & \text{\footnotesize if $C\cap\mathbb{I}^{j}\neq\emptyset, C\cap\mathbb{M}\neq\emptyset $,}
		\end{array} \right.  \label{eq:NewBernoulliProbEx}
\end{align}
\label{eq:UpdatedPGFLprevdet}%
\end{subequations}%
and the updated \gls{ppp} intensity is $D^{u}(\vv{x}) = (1 - p_{\rm D}) D_{+}^{u}(\vv{x})$. For Gaussian mixture intensity $D_{+}^u$, and Gaussian densities $f_{+}^{j,i}$, the updates $\ip{\cdot}{\phi}$ in \eqref{eq:UpdatedPGFLprevdet} are easily computed using the UKF update, see, e.g., \cite[Ch. 5]{Sarkka2013}.

\input{extraction}


%% file: extraction.tex
\subsubsection{Extraction}
\label{sec:ObjectExtraction}
Let the set of updated \gls{pmbm} parameters be
\begin{align}
	D_{}^{u} , \{ (w_{}^{j}, \{(r_{}^{j,i},f_{}^{j,i})\}_{i\in\mathbb{I}_{}^{j}})\}_{j\in\mathbb{J}_{}}.
\end{align}
To extract a set of object estimates, the hypothesis with highest probability is chosen,
\begin{equation}
    j^{\star}=\argmax_{j\in\idxs{J}}w^j.
 \end{equation}
 From the corresponding \gls{mb}, with parameters
 \begin{align}
	 \{(r_{}^{j^{\star},i},f_{}^{j^{\star},i})\}_{i\in\mathbb{I}_{}^{j^{\star}}},
 \end{align}
 all Bernoulli components with probability of existence $r^{j^{\star},i}$ larger than a threshold $\tau$ are selected, and the expected value of the object state is included in the set of object estimates,
 \begin{subequations}
 \begin{align}
 	\hat{\set{X}} & = \left\{ \hat{\vv{x}}^{j^{\star},i} \right\}_{i \in \mathbb{I}^{j^{\star}} : r^{j^{\star},i} > \tau} , \\
	\hat{\vv{x}}^{j^{\star},i} & = E_{f^{j^{\star},i}}\left[ \vv{x}^{j^{\star},i} \right] = \int \vv{x} f^{j^{\star},i}(\vv{x}) d\vv{x}.
 \end{align}
\end{subequations}



%% file: results.tex
\section{Experimental results}
\label{sec:results}

\subsection{Setup}
For evaluation, the KITTI object tracking dataset~\cite{Geiger2012CVPR} is used.
The datasets consists of 21 training sequences and 29 testing sequences that were collected using sensors mounted on a moving car.
Each sequence has been manually annotated with ground truth information, e.g., in the images, objects from the classes \textit{Car}, \textit{Pedestrian} and \textit{Cyclist} have been marked by bounding boxes.
In this work, the training dataset was split into two parts; one for training the \gls{cnn}, and one for validation.
The sequences used for training are 0, 2, 3, 4, 5, 7, 9, 11, 17 and 20, and the remaining ones are used for validation.

\subsection{Evaluation}
In this work we are primarily interested in the 3D tracking results. However, the KITTI testing sequences evaluate the tracking in 2D, hence we present results in both 2D and 3D. Performance is evaluated using \gls{iou} of the image plane bounding boxes and Euclidean distance as distance measurements, respectively.
For a valid correspondence between a \gls{gt} object and an estimated object, the 2D \gls{iou} has to be at least \SI{50}{\percent}, and the 3D Euclidean distance has to be within \SI{3}{\meter}, for the 2D and 3D evaluation, respectively.
The performance is evaluated using the CLEAR MOT performance measures~\cite{Bernardin2008}, including \gls{mota}, \gls{motp}, with addition of \gls{mt}, \gls{ml}, \gls{ids} and \gls{frag} from~\cite{YuanLi2009}, and \gls{f1}, \gls{pre}, \gls{rec} and \gls{far}.
The \glsentrylong{f1} is the weighted harmonic mean of the \glsentrylong{pre} and \glsentrylong{rec}.
Note that, for the 2D \gls{iou} measure, a larger value is better, whereas for the 3D Euclidean distance, lower is better.

\subsection{Results}

\begin{figure*}[!ht]
  \centering
  \subfloat[][]{
    \begin{tabular}[b]{c}
      \includegraphics[width=2.1in]{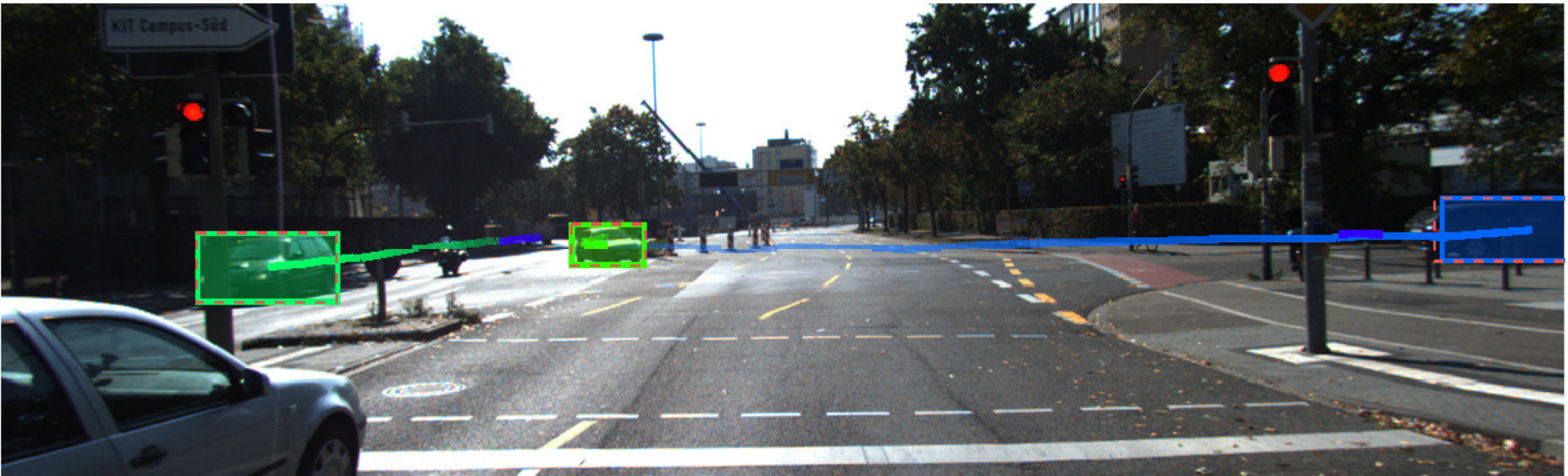}\\
      \includegraphics[width=2.1in]{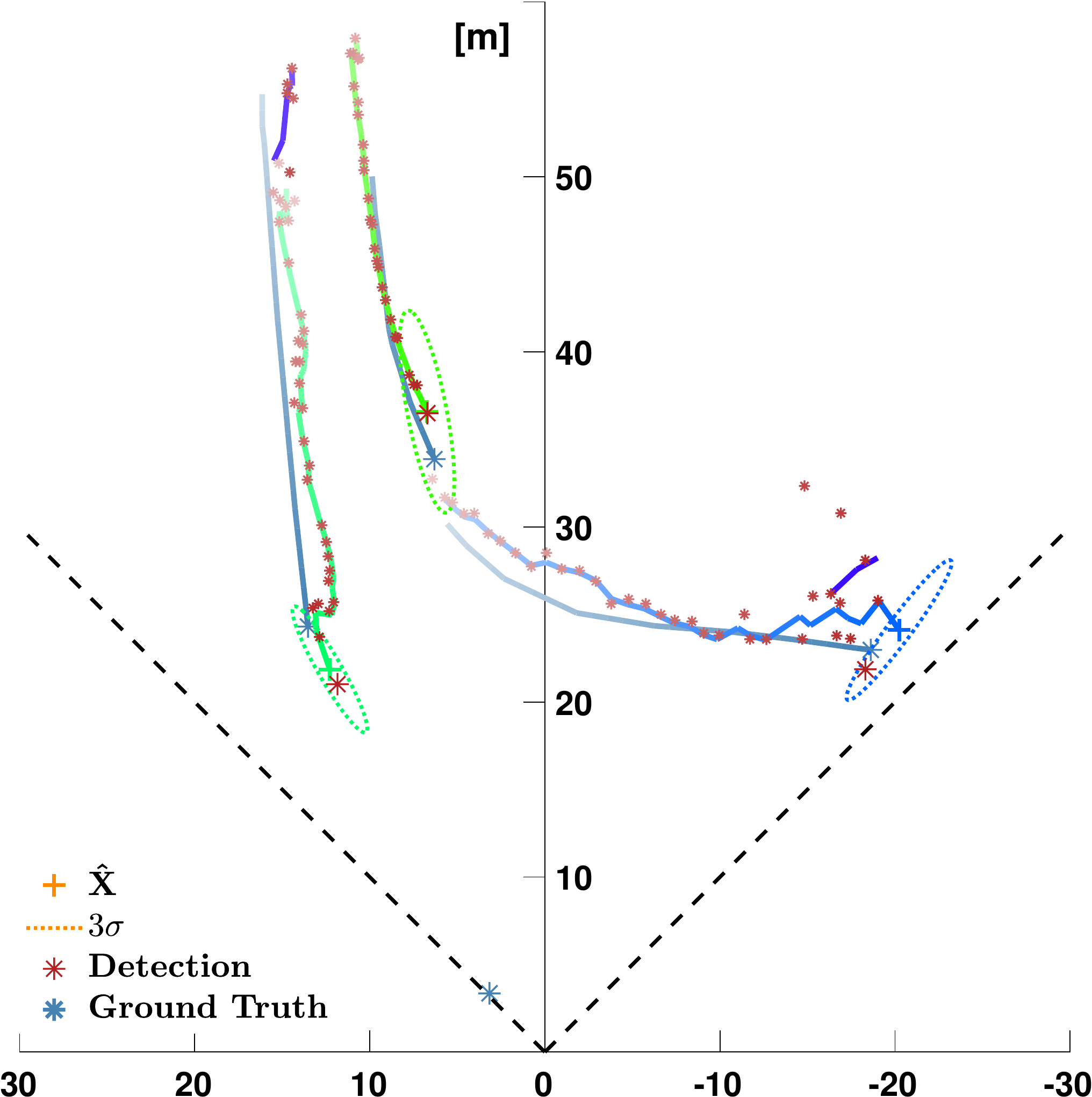}
    \end{tabular}
    \label{fig:qualitative_results_a}
  }
  \subfloat[][]{
    \begin{tabular}[b]{c}
      \includegraphics[width=2.1in]{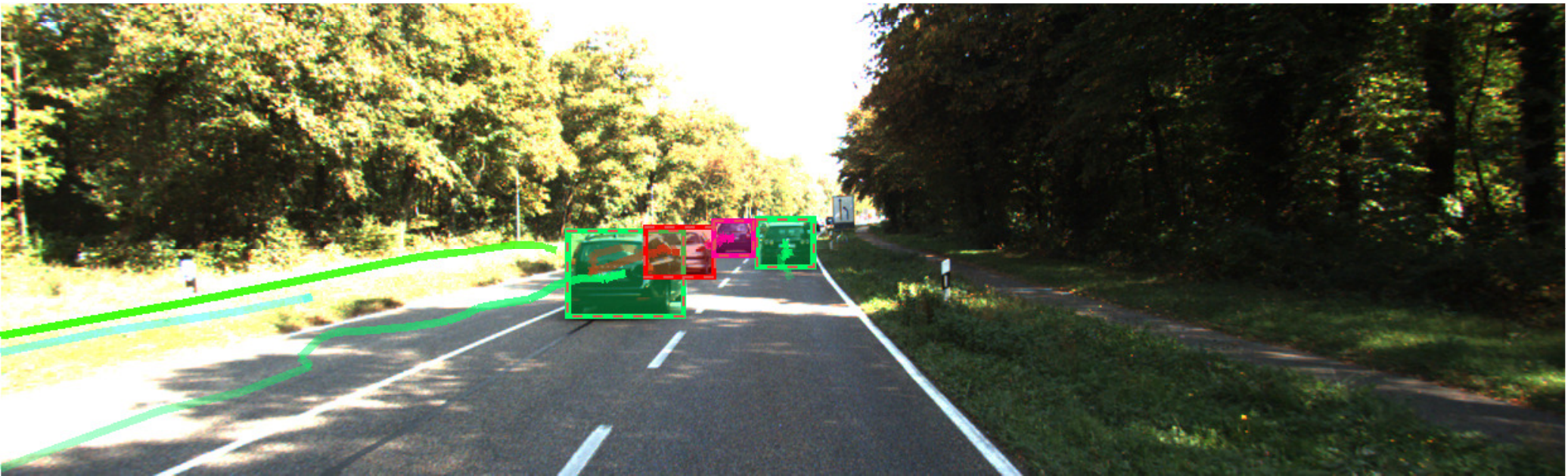}\\
      \includegraphics[width=2.1in]{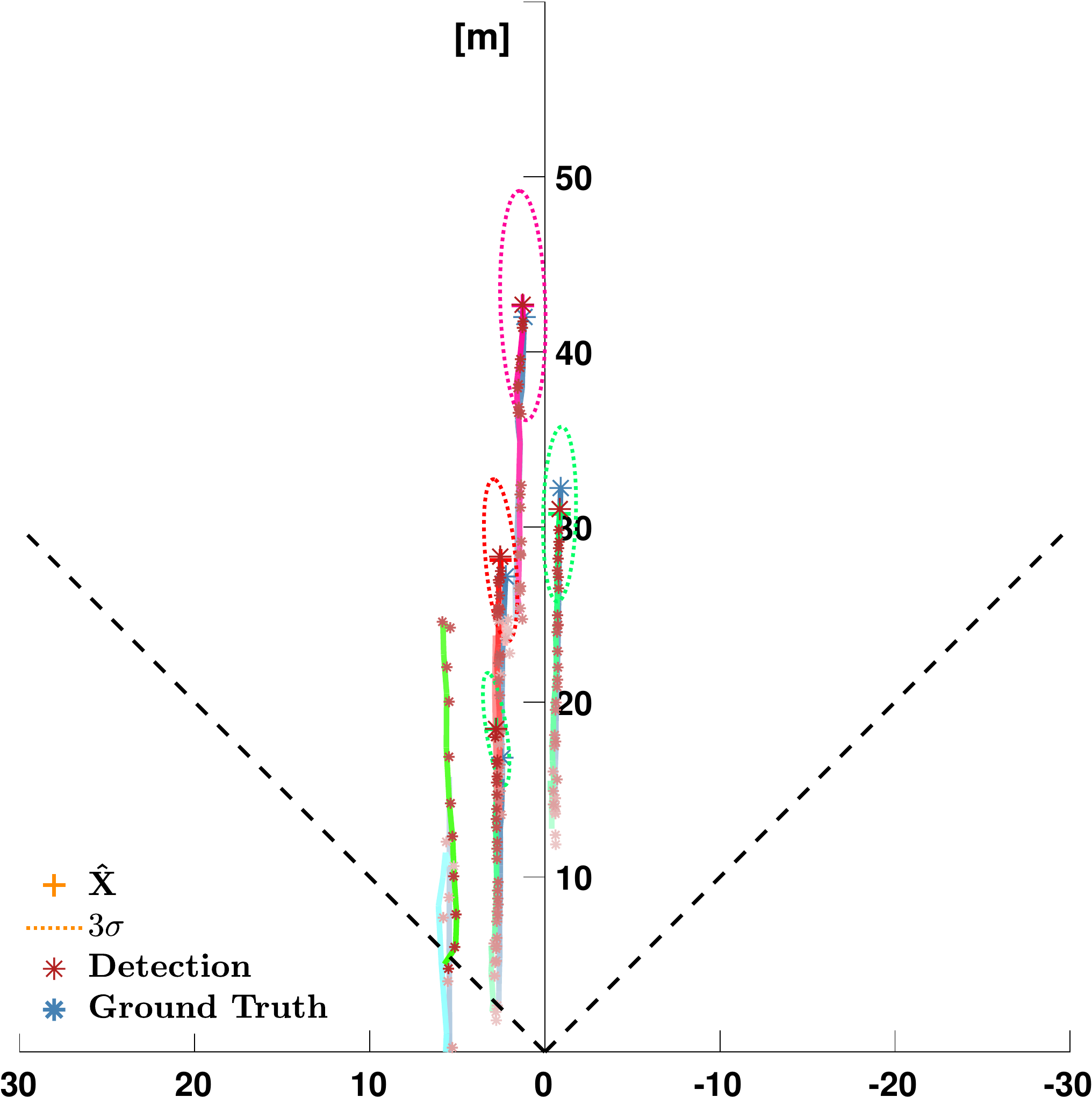}
    \end{tabular}
    \label{fig:qualitative_results_b}
  }
  \subfloat[][]{
    \begin{tabular}[b]{c}
      \includegraphics[width=2.1in]{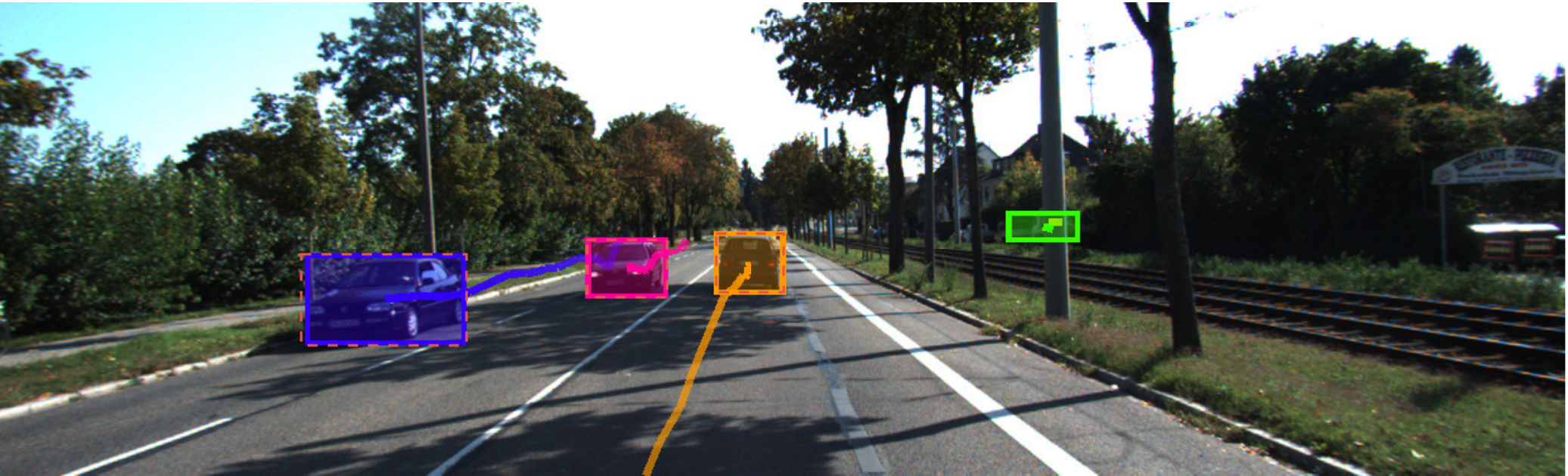}\\
      \includegraphics[width=2.1in]{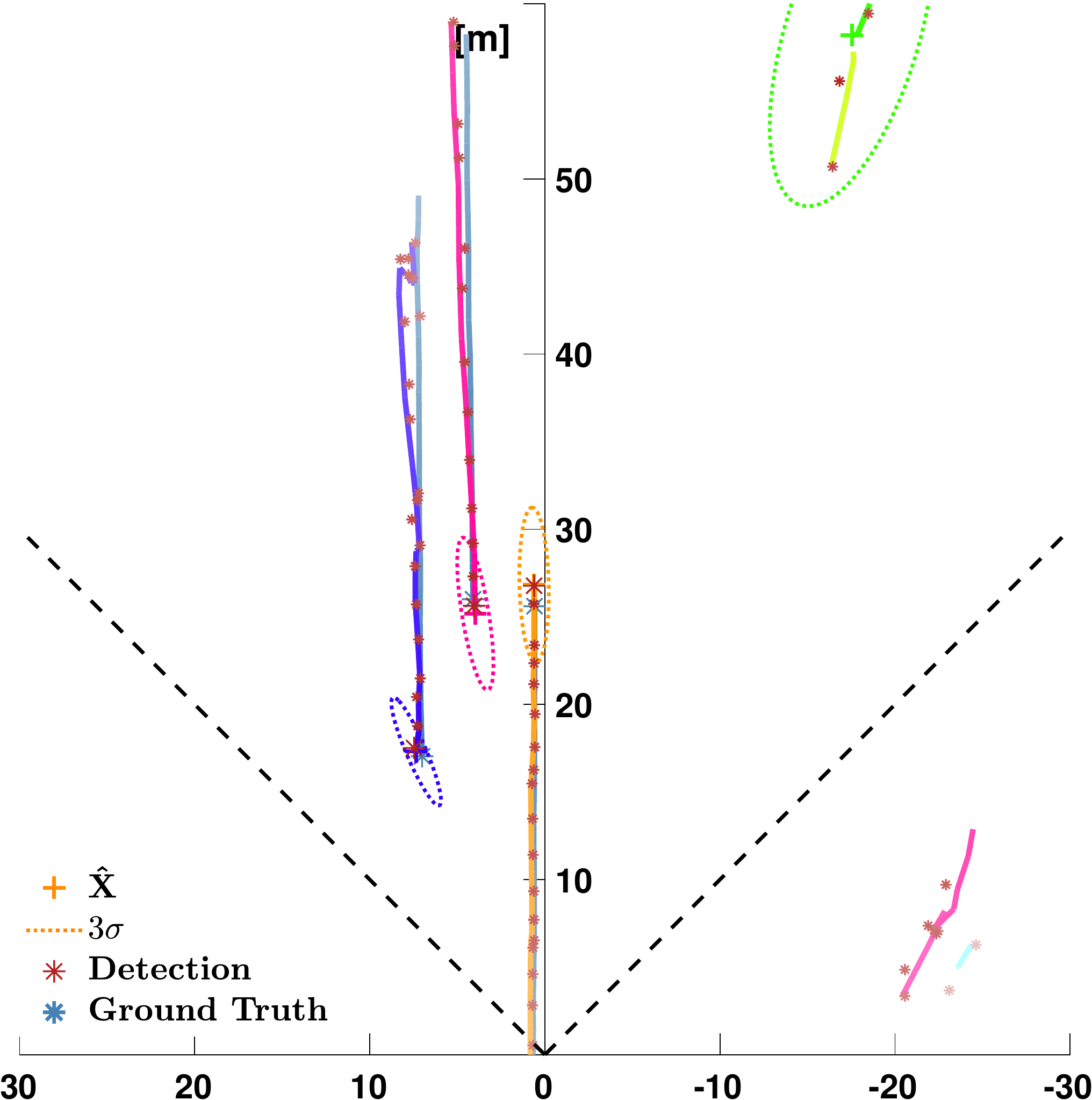}
    \end{tabular}
    \label{fig:qualitative_results_c}
  }
  \caption{Object estimates shown both projected to the image plane (top) and in top-view (bottom). The dotted tracks illustrate the \glsshort{gt} and tracked object estimates at previous time steps. The color of the tracks corresponds to the identity of the tracked object.}
  \label{fig:qualitative_results}
\end{figure*}

Examples of the 3D tracking results are shown in~\cref{fig:qualitative_results}.
The three examples show that the tracking algorithm successfully estimates the states of vehicles moving in the same direction as the ego-vehicle, vehicles moving in the opposite direction, as well as vehicles making sharp turns in intersections. In dense scenarios, such as in~\cref{fig:qualitative_results_b}, there are big overlaps between the bounding boxes; this is handled without problem by the data association. Noteworthy is that the distance estimates are quite noisy. Sometimes this leads to incorrect initial estimates of the velocity vector, as can be seen at the beginning of the track of the oncoming vehicle labelled purple in~\cref{fig:qualitative_results_c}. However, the tracking filter quickly converges to a correct estimate.
Videos of these, and of additional sequences, can be seen at \mbox{\url{https://goo.gl/AoydgW}}.

Quantitative results from the evaluation on the validation sequences are shown in~\cref{table:performance}. Noteworthy is the low amount of \glsentrylong{ids}, in both 2D and in 3D. Comparing the raw \gls{cnn} detections and the \gls{mot} algorithm, the \gls{mot} \glsentrylong{pre} is lower, and the \gls{mot} \glsentrylong{rec} is higher, leading to an \glsentrylong{f1} that is higher for the \gls{mot} than for the \gls{cnn}; in other words, the overall object detection performance is slightly improved.

\begin{table*}[ht]
  \centering
  \caption{Results of the evaluation of tracking performance in 2D and 3D on the \textit{Car} class. $\uparrow$ and $\downarrow$ represents that high values and low values are better, respectively. The best values are marked with bold font.}
  \label{table:performance}

  \begin{tabular}{ll|rrrrrrrrrrr}
     & Method & \glsshort{mota}$\uparrow$ & \glsshort{motp}$\uparrow$ & \glsshort{mt}$\uparrow$ & \glsshort{ml}$\downarrow$ & \glsshort{ids}$\downarrow$ & \glsshort{frag}$\downarrow$ & \glsshort{f1}$\uparrow$ & \glsshort{pre}$\uparrow$ & \glsshort{rec}$\uparrow$ & \glsshort{far}$\downarrow$ \\ \hline
    \multirow{2}{*}{2D}
     & \glsshort{cnn} &
                        -- & {\bf \SI{82.04}{\percent}} & \SI{74.59}{\percent} & \SI{3.78}{\percent} & -- & -- & \SI{91.16}{\percent} & {\bf \SI{95.72}{\percent}} & \SI{87.02}{\percent} & {\bf \SI{9.08}{\percent}}
    \\
     & \glsshort{mot}  &
                         {\bf \SI{81.23}{\percent}} & \SI{81.63}{\percent} & {\bf \SI{76.22}{\percent}} & \SI{3.78}{\percent} & 19 & 107 & {\bf \SI{91.26}{\percent}} & \SI{94.76}{\percent} & {\bf \SI{88.02}{\percent}} & \SI{11.46}{\percent}
    \\ \hline
    & Method & \glsshort{mota}$\uparrow$ & \glsshort{motp}$\downarrow$ & \glsshort{mt}$\uparrow$ & \glsshort{ml}$\downarrow$ & \glsshort{ids}$\downarrow$ & \glsshort{frag}$\downarrow$ & \glsshort{f1}$\uparrow$ & \glsshort{pre}$\uparrow$ & \glsshort{rec}$\uparrow$ & \glsshort{far}$\downarrow$ \\ \hline
    \multirow{2}{*}{3D}
     & \glsshort{cnn} &
                        -- & \SI{111.39}{\centi\meter} & \SI{45.95}{\percent} & {\bf \SI{10.27}{\percent}} & -- & -- & \SI{73.53}{\percent} & {\bf \SI{78.74}{\percent}} & \SI{68.97}{\percent} & {\bf \SI{41.90}{\percent}}
    \\
     & \glsshort{mot}  &
                         {\bf \SI{47.20}{\percent}} & {\bf \SI{110.73}{\centi\meter}} & {\bf \SI{48.65}{\percent}} & \SI{11.35}{\percent} & 20 & 166 & {\bf \SI{73.86}{\percent}} & \SI{78.18}{\percent} & {\bf \SI{70.00}{\percent}} & \SI{44.32}{\percent}
  \end{tabular}
\end{table*}



The runtime of the algorithm is on average in total \SI{52}{\milli\second}, \SI{38}{\milli\second} for the detection network and \SI{14}{\milli\second} for the tracking algorithm, on a Nvidia Tesla V100 SXM2 and a single thread on a \SI{2.7}{\giga\hertz} Intel Core i7.

\subsection{KITTI \textsc{\glsentryshort{mot}} benchmark}
The \gls{mot} algorithm was also evaluated in 2D using the test sequences on the KITTI evaluation server.
For these results, the full training set was used for training the detection \gls{cnn}. The results are presented in~\cref{table:kitti_mot_performance}; at the time of submission our algorithm was ranked \nth{3} in terms of \gls{mota} among the published algorithms.
Note that, even if not reaching the same \gls{mota} performance, the runtime of our algorithm (\gls{fps}) is one magnitude faster and has a significantly lower number of \glsentrylong{ids} than the two algorithms with higher \gls{mota}.

\begin{table}[ht]
  \begin{threeparttable}
  \centering
  \caption{KITTI \glsentryshort{mot} benchmark~\cite{Geiger2012CVPR} results for \textit{Car} class. $\uparrow$ and $\downarrow$ represents that high values and low values are better, respectively. The best values are marked with bold font. \cite{Choi2016} is TuSimple, \cite{Xiang2015} is IMMDP and \cite{Lee2016} is MCMOT-CPD. Only results of published methods are reported.}
  \label{table:kitti_mot_performance}
  \begin{tabular}{l | r | r | r | r | r | r | r}
    & \glsshort{mota}$\uparrow$ & \glsshort{motp}$\uparrow$ & \glsshort{mt}$\uparrow$ & \glsshort{ml}$\downarrow$ & \glsshort{ids}$\downarrow$ & \glsshort{frag}$\downarrow$ & \glsshort{fps}$\uparrow$\tnote{a}
    \\ \hline
    \cite{Choi2016} & {\bf\SI{86.6}{\percent}} & {\bf\SI{84.0}{\percent}} & {\bf\SI{72.5}{\percent}} & \SI{6.8}{\percent} & 293 & 501 & 1.7 \\
    \cite{Xiang2015} & \SI{83.0}{\percent} & \SI{82.7}{\percent} & \SI{60.6}{\percent} & \SI{11.4}{\percent} & 172 & {\bf 365} & 5.3 \\
    Our & \SI{80.4}{\percent} & \SI{81.3}{\percent} & \SI{62.8}{\percent} & {\bf\SI{6.2}{\percent}} & {\bf 121} & 613 & 73 \\
    \cite{Lee2016} & \SI{78.9}{\percent} & \SI{82.1}{\percent} & \SI{52.3}{\percent} & \SI{11.7}{\percent} & 228 & 536 & {\bf 100} \\
  \end{tabular}
  \begin{tablenotes}
      \footnotesize
      \item[a] The time for object detection is not included in the specified runtime.
    \end{tablenotes}
\end{threeparttable}
\end{table}


%% file: conclusions.tex
\section{Conclusion}
\label{sec:Conclusion}
This paper presented an image based \gls{mot} algorithm using deep learning detections and \gls{pmbm} filtering.
It was shown that a \gls{cnn} and a subsequent \gls{pmbm} filter can be used to detect and track objects.
The algorithm successfully can track multiple objects in 3D from a single camera image, which can provide valuable information for decision making and control.


%% file: acknowledgment.tex
\section*{Acknowledgment}
This work was partially supported by the Wallenberg Autonomous Systems and Software Program (WASP).
